\documentclass[conference,a4paper]{IEEEtran}
\IEEEoverridecommandlockouts

\usepackage[hidelinks]{hyperref}
\usepackage[cmex10]{amsmath}
\usepackage{amssymb,amsfonts}
\usepackage{dblfloatfix}

\usepackage[ruled,vlined]{algorithm2e}
\usepackage{graphicx}
\graphicspath{{Figures/PDF/}{Figures/PNG/}}

\usepackage{booktabs}
\usepackage{siunitx}
\usepackage[numbers,compress]{natbib}
\usepackage{texnames}
\usepackage{bm,bbm}
\usepackage{orcidlink}

\begin{document}

\title{\uppercase{STCAD: Scalable Trajectory Clustering and Anomaly Detection on Terabyte-Scale AIS Data}
}

\author{
	\IEEEauthorblockN{
        Bertram Hage\orcidlink{0009-0005-5078-8363},
        Alexander Schiøtz\orcidlink{0009-0002-1144-9885},
        Felix Thomsen\orcidlink{0009-0009-8857-9267},
        Christian Rand\orcidlink{0000-0002-0574-0783},
        Peder Heiselberg\orcidlink{0000-0002-8847-634X}
        \thanks{
        \copyright{} 2026 IEEE.  Personal use of this material is permitted.  Permission from IEEE must be obtained for all other uses, in any current or future media, including reprinting/republishing this material for advertising or promotional purposes, creating new collective works, for resale or redistribution to servers or lists, or reuse of any copyrighted component of this work in other works.
        }
        }
\IEEEauthorblockA{
\textit{Technical University of Denmark}\\
Anker Engelunds Vej 101
2800 Kongens Lyngby, Denmark\\
ph@space.dtu.dk
}
}

\maketitle
\begin{abstract}
We present a scalable framework for unsupervised clustering of maritime trajectories derived from terabyte-scale Automatic Identification System (AIS) archives. Variable-length trajectories are encoded with a custom BERT-based model trained via masked token modeling and clustered using CURE hierarchical clustering, producing physically interpretable trajectory groups without requiring a predefined number of clusters. An intrinsic unsupervised anomaly detection method based on reconstruction loss and clustering noise assignment identifies irregular navigation patterns. The framework is demonstrated on a national-scale AIS dataset comprising billions of messages spanning one year, yielding stable trajectory clusters and a clear separation between nominal and anomalous vessel behavior.
\end{abstract}


\section{Introduction}
\label{sec:intro}
Ensuring maritime safety requires the ability to analyze extremely large and noisy AIS datasets to detect irregular vessel behaviors and potential risks. Maritime domain awareness relies heavily on data from the Automatic Identification System (AIS), which provides continuous, large-scale tracking of vessels. AIS is mandated for most commercial vessels and is widely used for collision avoidance, traffic monitoring, and maritime safety enforcement. As a result, national and regional AIS archives now contain billions of messages collected over long time periods \cite{DMA_AIS_Data}. These data consist of irregularly sampled and often noisy observations that, when aggregated, form variable-length vessel trajectories. The combination of scale, heterogeneity, and data quality issues poses substantial challenges.

Anomaly detection is typically formulated through clustering-based outlier identification or as the detection of deviations from learned normal behavior. 
Early frameworks focused on vessel pattern extraction and anomaly detection \cite{pallotta2013vessel}. Subsequent studies emphasized clustering-based approaches for their robustness to noise and irregular sampling \cite{han2021modeling, wang2021ship, wang2025application}, often combined with anomaly detection to identify irregular navigation patterns \cite{zhang2023novel}.
Transformer architectures are able to capture long-range dependencies in variable-length sequences and have therefore become a standard in AIS trajectory modeling \cite{transformers, advancing_ais, yang2024harnessing, traisformer, multi-path_trans, sorensen2022probabilistic}.
The Bidirectional Encoder Representations from Transformers (BERT) architecture \cite{bert_paper, roberta_paper} demonstrated that, with sufficient training, encoders can learn rich semantic representations of sequences solely through Masked Token Modeling (MTM). In this self-supervised approach, the model learns the underlying structure of a sequence by predicting randomly masked tokens within the input. 
In recent research, these unsupervised techniques have been applied to the maritime domain for trajectory pattern extraction \cite{karatacs2021trajectory, fujino2023navigation} and the identification of behavioral irregularities \cite{liang2024unsupervised}.

In this work, we present STCAD, a scalable unsupervised framework for trajectory clustering and anomaly detection on terabyte-scale AIS data. We combine a self-supervised BERT-style Transformer encoder \cite{bert_paper, roberta_paper} trained via masked token modeling with the CURE hierarchical clustering algorithm \cite{cure}. 
The framework is supported by a preprocessing pipeline, enabling year-scale analysis of billions of AIS messages. 
STCAD provides a unified framework for unsupervised anomaly detection on extremely large AIS datasets implemented using standard Python libraries \cite{numpy, pandas, pytorch, umap, hugging_face_transformers, polars}.

\section{Data}
\label{sec:data}
This study uses a national-scale AIS dataset provided by the Danish Maritime Authority (DMA) \cite{DMA_AIS_Data} gathered via their AIS network covering the Danish waters. This study uses data spanning the entire 2024 calendar year. This dataset has approximately 6.89 billion messages totaling 1.2 TB of raw data. This scale represents one of the largest temporal and spatial AIS datasets utilized in current maritime anomaly detection literature.
Each observation in the dataset corresponds to a single AIS message consisting of the Maritime Mobile Service Identity (MMSI) number vessel identifier, latitude, longitude, Speed over Ground (SOG) in knots, and Course over Ground (COG) in degrees.
While the DMA archive contains auxiliary static and voyage-related features, these often exhibit significant sparsity, particularly among smaller vessel classes, and were therefore excluded them.
The dataset consists of 38,026 unique vessels, of which 19,597 are sailing or pleasure crafts, 8,627 are cargo or tanker vessels, 1,632 are fishing vessels, and 8,170 are another category or undefined.

\subsection{Data preprocessing}\label{sec:preprocessing}
The raw AIS messages were converted into regularly sampled, valid, individual voyages in two phases: a global phase and a per-vessel (MMSI) phase.

First, we applied global filters to remove erroneous messages outside the geographical region of Denmark ($\text{lat} \in [54, 59]$, $\text{lon} \in [5, 17]$) and messages with unrealistic values (SOG $> 30$ knots or COG outside $[0, 360]$). This resulted in a dataset of 5,383,009,610 messages.

The per-vessel phase involved computationally intensive tasks, with initial experiments showing that processing a single MMSI could require up to $\sim$1GB of memory. This resulted in an estimated sequential runtime of $\sim$170 hours, or more than 7 days. To mitigate this, we implemented a distributed processing pipeline by mapping the data by MMSI and saving it as serialized pickle files in vessel-specific directories. This created key-value pairs of the form:
\[
\langle \text{MMSI}, \{m_{n,1}, m_{n,2}, \dots, m_{n,T}\} \rangle,
\]
where $m_{n,t}$ represents the $t$-th message for an MMSI on day $n \in \{1, 2, \dots, 366\}$.

The subsequent reduction phase, performed in parallel using the DTU Computing Center (DCC) \cite{dtu_hpc}, applied vessel-specific preprocessing rules. 
We defined a voyage as a contiguous trajectory where the interval between consecutive messages is less than 2 hours. Voyages exceeding 20 hours were split into shorter segments, while those shorter than 4 hours or containing fewer than 20 messages were discarded. To ensure robustness against outliers, we removed messages where the empirical speed (calculated via distance and timestamp) exceeded 40 knots. Finally, we applied linear interpolation to enforce a regular 5-minute sampling interval and performed min-max normalization on each feature.

By utilizing 64 CPU cores with 4GB of memory each (128 threads in total), we reduced the processing time to $\sim$80 minutes. Ultimately, these preprocessing steps reduced the dataset from 1,342,406 initial trajectories to 453,712 final voyages while staying below a feasible memory usage.

\section{Methodology}\label{sec:methodology}
This section describes the transformation of variable-length trajectory sequences into fixed-size embeddings and the subsequent clustering to identify distinct behaviors and anomalies.

\subsection{Sequence encoding}\label{sec:transformer}
Variable-length vessel trajectories are encoded into fixed-dimensional representations using a Transformer with a stack of self-attention layers capturing global dependencies. A classification token (\texttt{[CLS]}) is prepended to each sequence, and its final hidden state serves as the trajectory embedding.

Given the homogeneous structure of AIS data, compared to natural language, a reduced Transformer is employed with four encoder layers, four attention heads per layer, and a hidden size of 256, in contrast to larger architectures such as BERT-base \cite{bert_paper}. Training uses masked token modeling with a 15\% masking probability, and dropout of 0.1 is applied to both hidden states and attention probabilities.

Each AIS message is represented by the spatial and kinematic features:
\[
[\text{lat}, \text{lon}, \text{SOG}, \text{COG}] \in \mathbb{R}^4
\]
which are projected into the model hidden space using a linear mapping,
\[
f: \mathbb{R}^4 \rightarrow \mathbb{R}^{256},
\]
after which positional encodings are added to preserve temporal ordering. The \texttt{[CLS]} token is represented by the fixed vector $[-1.0, -1.0, -1.0, -1.0]^\top$ prior to projection and is processed identically to all other tokens during encoding.

\subsection{Hierarchical clustering with CURE}
Agglomerative Clustering (AC) is powerful because it builds hierarchies without requiring a preset number of clusters, allowing us to characterize stable trajectory cluster sizes in our embeddings. However, standard AC is computationally prohibitive for large datasets due to its $O(n^3)$ time complexity. To address this, we employ Clustering Using REpresentatives (CURE) \cite{cure}. By clustering a sample ($N=1000$) and assigning the remaining points to the nearest representative set, CURE reduces the fitting time to linear $O(n)$ complexity. Prior to clustering, the embeddings are L2-normalized, ensuring that the Euclidean distance used by the algorithm corresponds to the cosine similarity, the standard metric for assessing semantic similarity.

We define an \textit{assignment threshold} where a point is labelled as noise if the distance to the nearest representative of a cluster is greater than the threshold. 
To avoid accidentally including a noisy point as part of our representatives, we discard samples from the AC sample set that remain isolated after the first $N \cdot \rho$ merges, where we set $\rho=0.05$.

\section{Experiments and results}
We train our encoder model described in \ref{sec:transformer} on 80\% of our preprocessed dataset, keeping a 20\% hold out validation set. The model reached a stable validation MSE loss of $\sim 0.0044$ after 17 epochs of training on an A100 Nvidia GPU with batch size 512 and the Adam optimizer \cite{loshchilov2017decoupled} with learning rate $5\cdot 10^{-5}$.

\begin{figure}
    \centering
    \includegraphics[width=\linewidth]{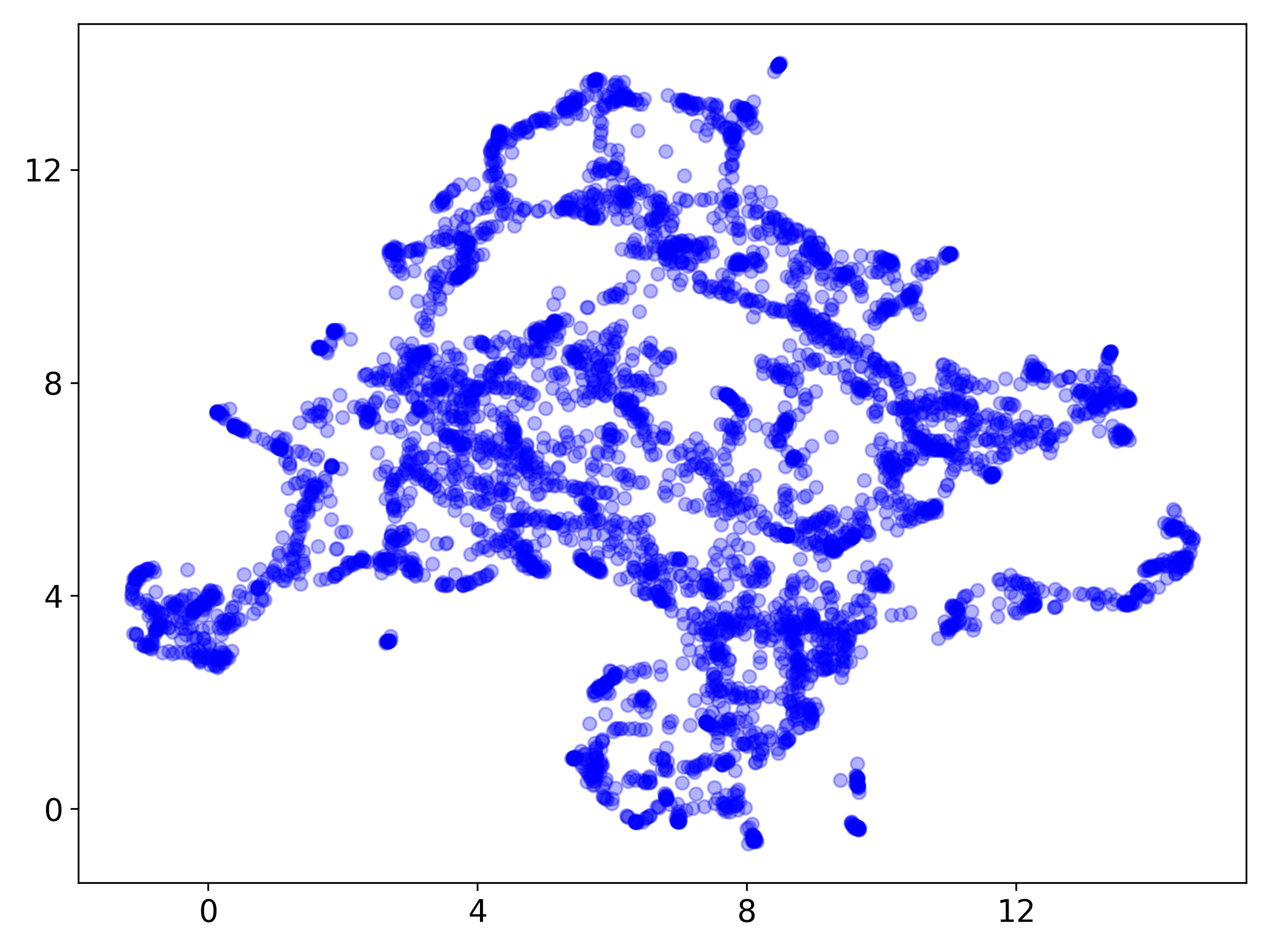}
    \caption{2D UMAP visualisation of the embeddings.}
    \label{fig:umap}
\end{figure}

The two-dimensional Uniform Manifold Approximation and Projection (UMAP) \cite{umap} visualization of the embedding space (Figure~\ref{fig:umap}) exhibits clusters with non-spherical geometries. This characteristic motivates the consideration of hierarchical clustering approaches that do not impose spherical cluster assumptions.

Single linkage clustering \cite{sibson1973slink} was evaluated but exhibited pronounced chaining effects, whereby isolated noise points induced premature cluster merging. Average linkage \cite{seifoddini1989single} produced a highly fragmented and unstable dendrogram, limiting interpretability and robustness.

Despite the apparent non-spherical structure suggested by the UMAP projection, agglomerative clustering with Within-Cluster Sum of Squares (WARD) linkage \cite{ward1963hierarchical} yielded the most coherent and stable cluster partitions and was therefore adopted in the subsequent analysis.

\begin{figure}
    \centering
    \includegraphics[width=\linewidth]{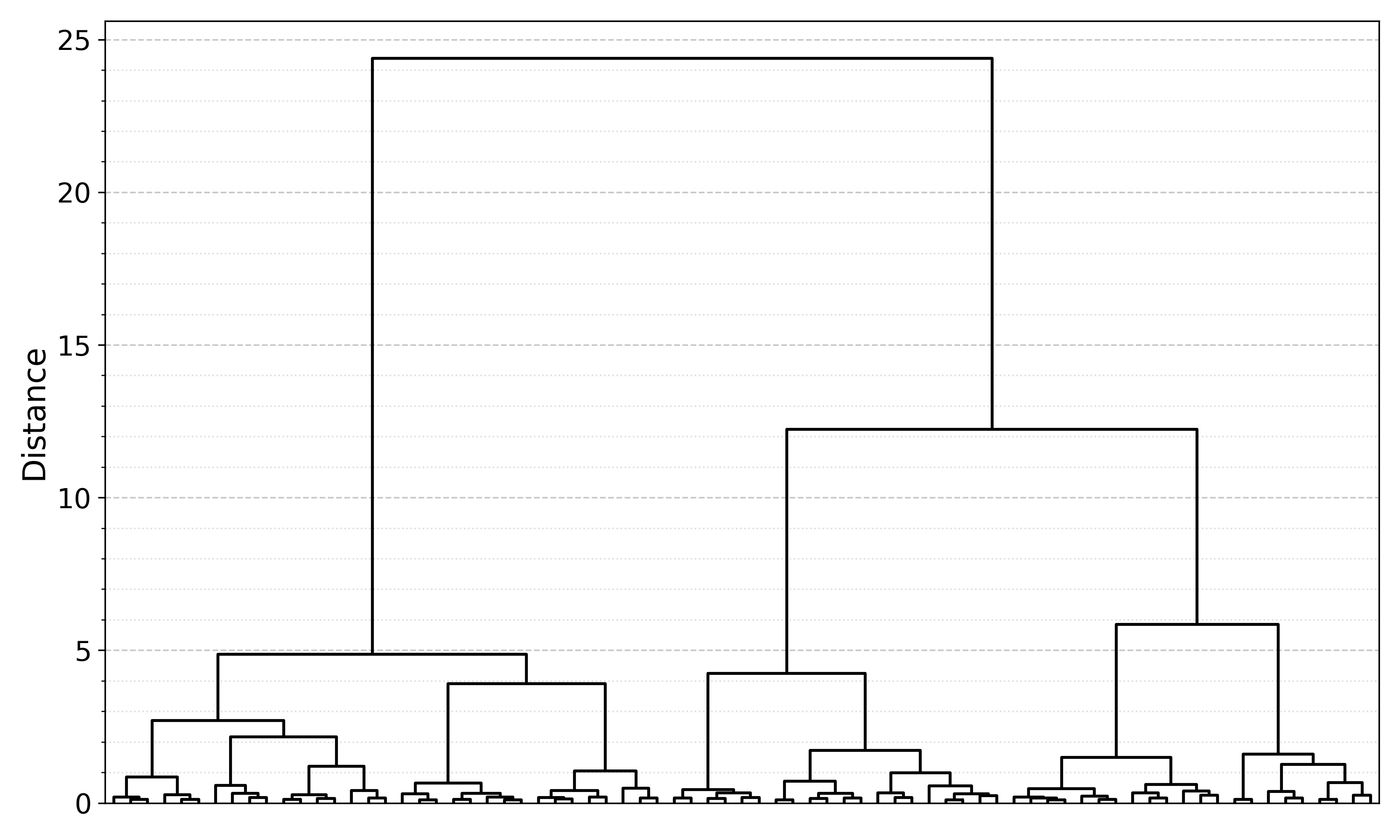}
    \caption{Dendrogram from the Agglomerative Clustering with WARD linkage for 1000 samples. Truncated to the last 75 merges.}
    \label{fig:dendrogram}
\end{figure}

CURE hyperparameters were selected based on preliminary experiments, with the compression factor set to 0.6 and the number of representative points set to 20. Inspection of the dendrogram in Figure~\ref{fig:dendrogram} indicates stable partitioning at cluster counts $k \in \{2, 3, 5, 8\}$. Subsequent analyses are conducted for $k = 3$ and $k = 5$, which provide a balance between partition stability and analytical granularity.

\begin{figure}[t]
    \centering
    \includegraphics[width=1\linewidth]{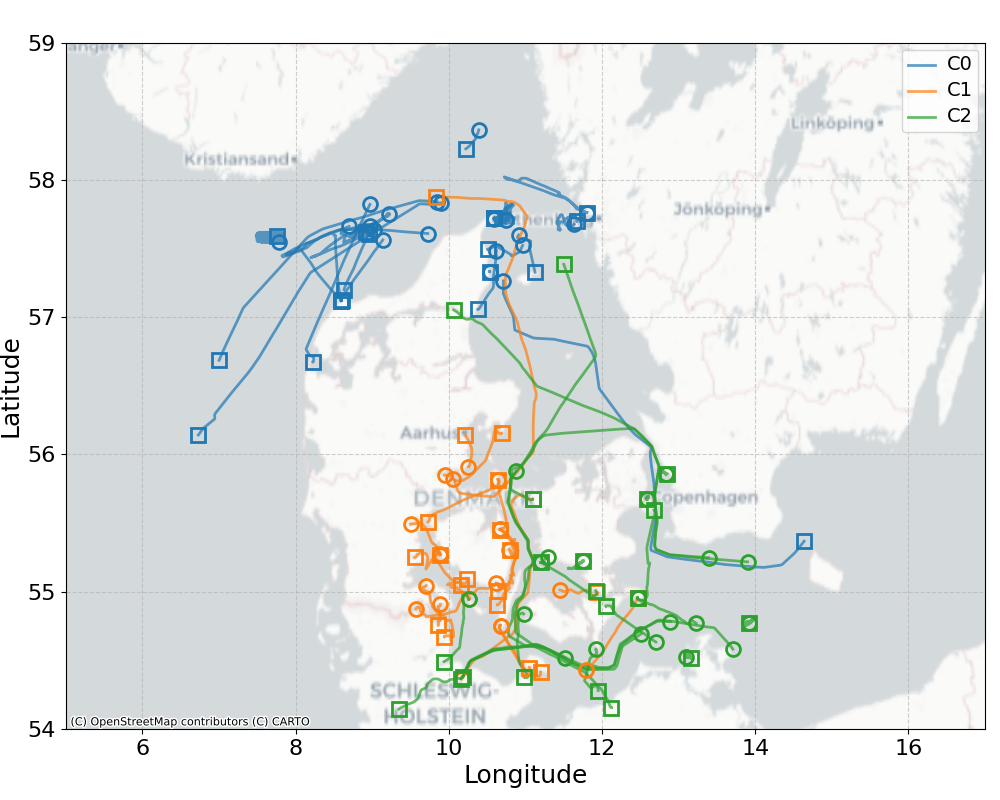}                 \centering
       \includegraphics[width=1\linewidth]{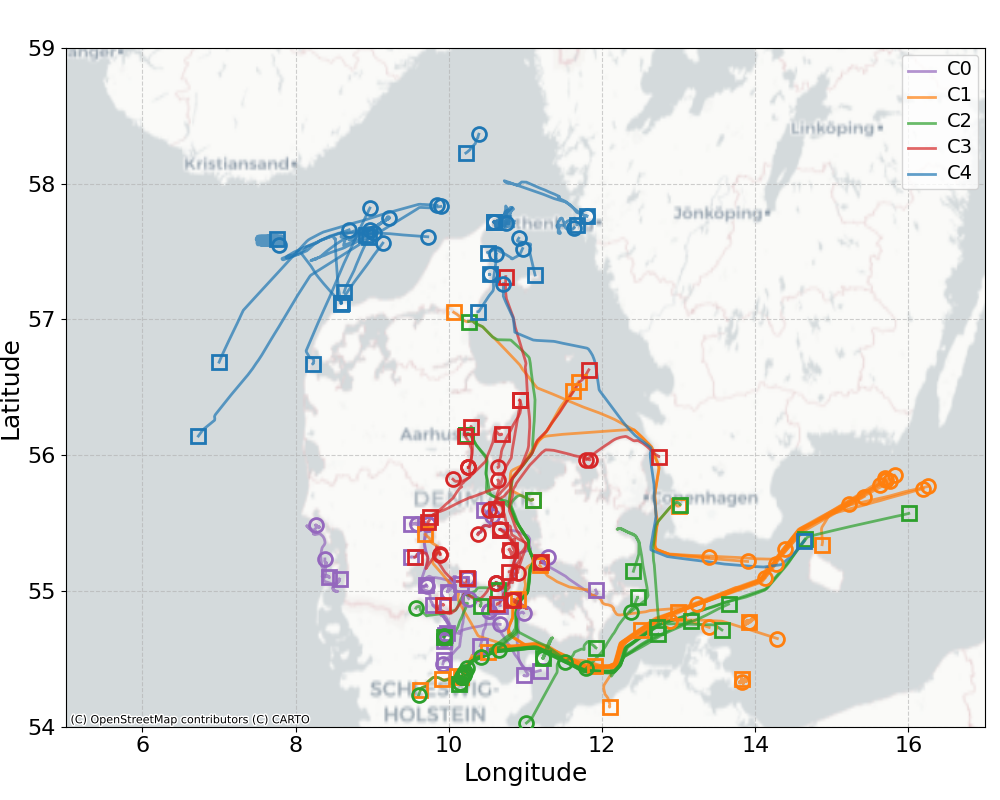}
      \caption{The 20 trajectories closest to each cluster centroid, out of the samples used for agglomerative clustering, with 3 (a) and 5 (b) clusters. $\bigcirc$ and $\square$ marks the start and end of trajectories, respectively.}
    \label{fig:traj}

\end{figure}

Representative trajectories for these cluster resolutions are shown in Figure~\ref{fig:traj}. At $k = 3$, one cluster is primarily associated with trajectories located north of the Jutland Peninsula. Increasing the resolution to $k = 5$ (Figure~\ref{fig:traj} bottom) further subdivides two of the clusters, resulting in more spatially localized trajectory groups with reduced spatial extent.

Cluster characterization is performed using the following trajectory-level descriptors:
\begin{itemize}
    \item Mean latitude and longitude $(\textbf{lat}, \textbf{lon})$
    \item Mean speed over ground ($\textbf{Speed}$)
    \item Start-to-end displacement in meters (\textbf{Disp.})
    \item Course variability $\sigma_{COG} (\textbf{Turn})$
    \item Vessel type (\textbf{Cargo}, \textbf{Tanker}, \textbf{Fishing}, \textbf{Sailing/Pleasure}, \textbf{Other}) 
\end{itemize}

We use the Z-score to assess deviations in demographic features for the continuous features.
\begin{equation}\label{eq:zscore}
Z = \frac{\mu_{\text{cluster},\text{feature}} - \mu_{\text{global},\text{feature}}}{\sigma_{\text{global},\text{feature}}}.
\end{equation}
and Pointwise Mutual Information ($PMI$) score for the categorical vessel types
\begin{equation}\label{eq:pmi}
PMI_{V=v} = \log_2\left(\frac{P(V=v|\text{cluster})}{P(V=v)}\right),
\end{equation}
where $V$ represents the vessel type feature and $v$ denotes a specific category. The $\log_2$ in \eqref{eq:pmi} serves to symmetrize deviations in the negative and positive directions. 

\begin{figure}[t]
    \centering
        \includegraphics[width=\linewidth]{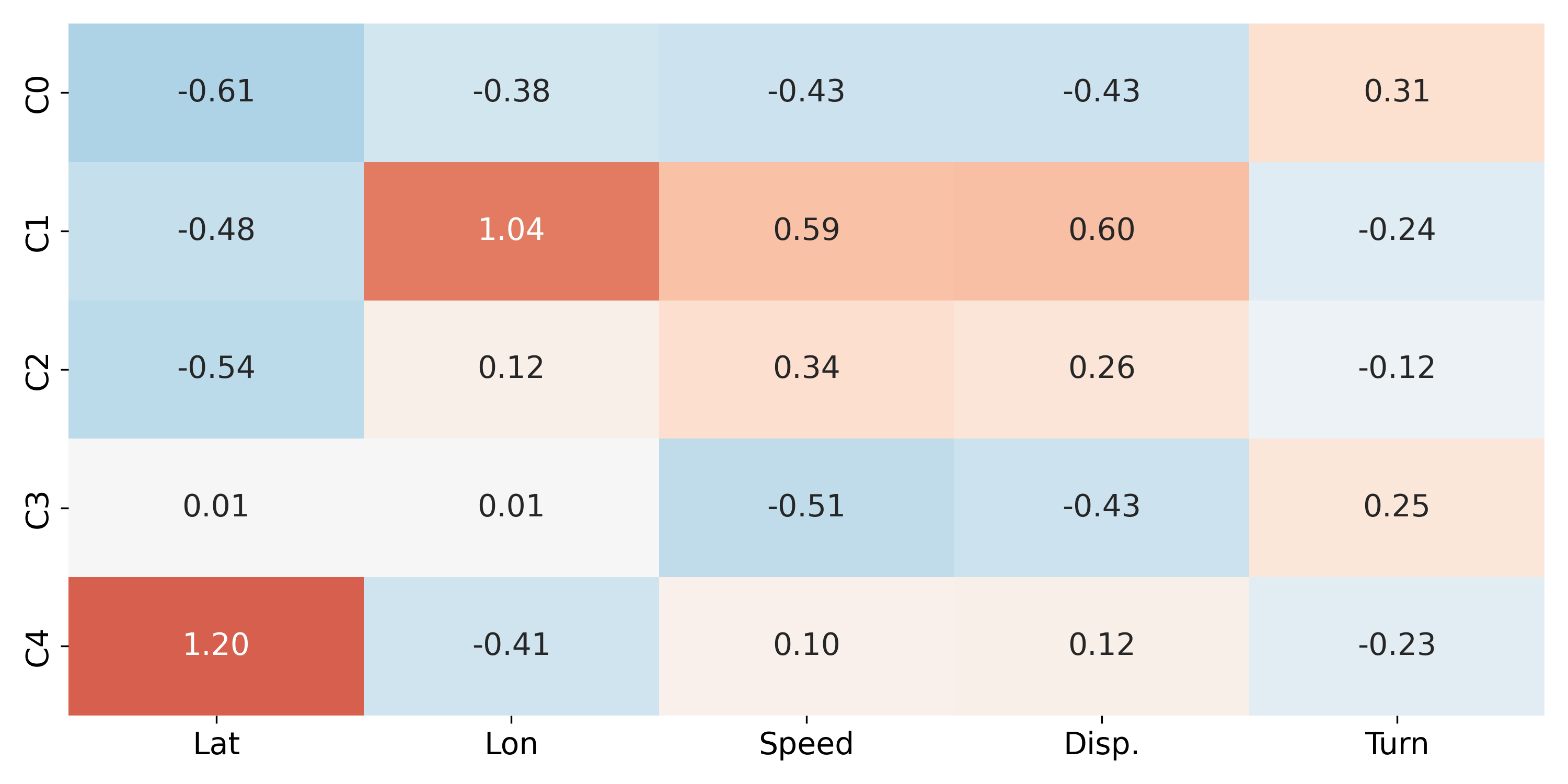} 
        \centering
        \includegraphics[width=\linewidth]{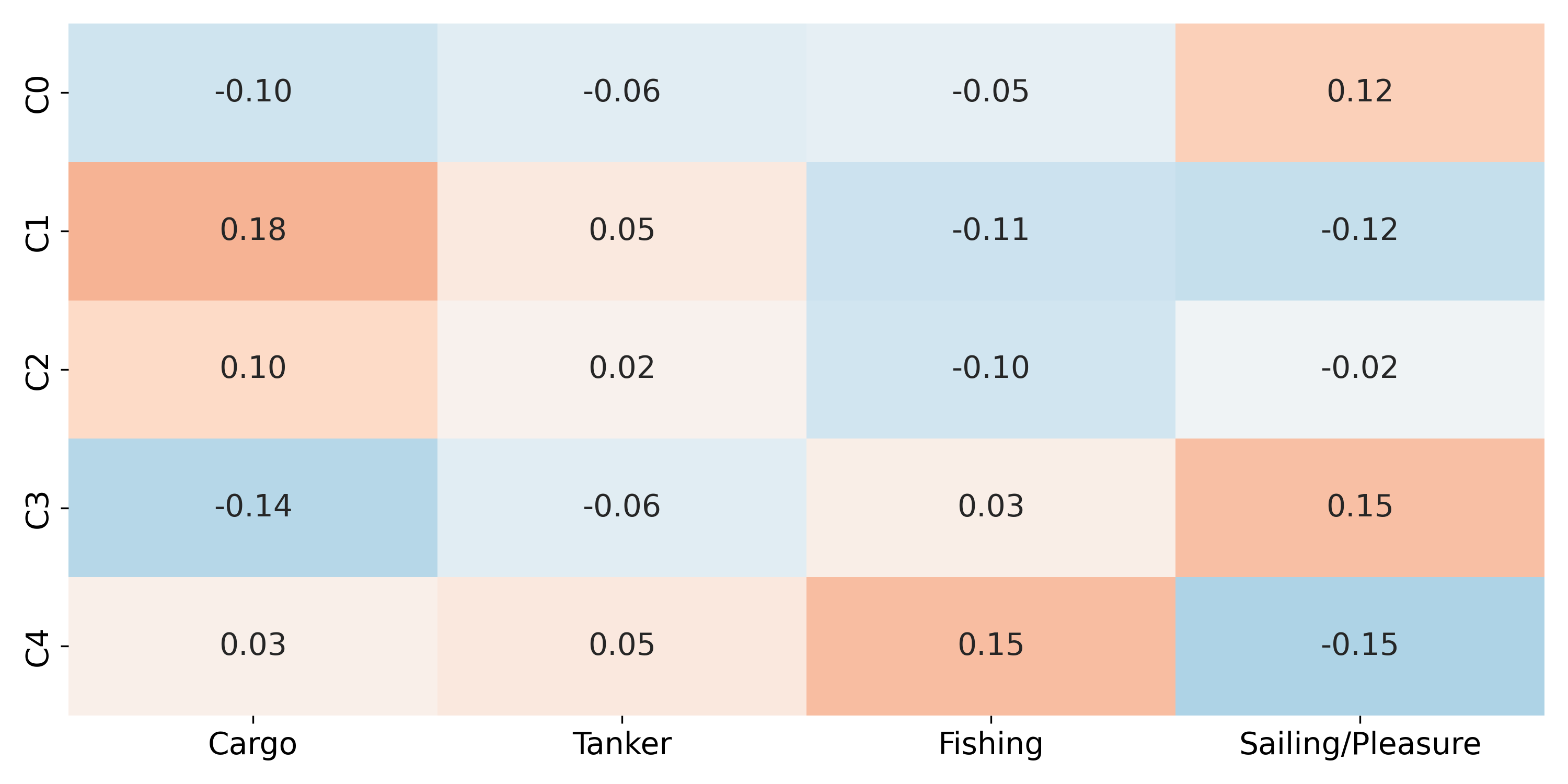}    
    \caption{Deviations in physical features for 5 clusters (C0-C4). (Top) Z-score for continuous features and (Bottom) $PMI$ for vessel types.}
    \label{fig:features}
\end{figure}
The cluster descriptors reveal distinct spatial, kinematic, and vessel-type patterns (see Figure~\ref{fig:features}). Cluster~4 (C4) is characterized by high latitudes and low longitudes, consistent with the spatial patterns in Figure~\ref{fig:traj}, and shows a predominance of fishing vessels. Clusters~1 and~2 (C1, C2) are dominated by cargo vessels with high speed and displacement, corresponding to transit through the Kiel Canal and onward routing toward Danish harbors. In contrast, clusters~0 and~3 (C0, C3) primarily consist of low-speed, low-displacement pleasure crafts spatially concentrated south of Jutland and around the island of Funen, respectively.

\begin{figure}
    \centering
    \includegraphics[width=1\linewidth]{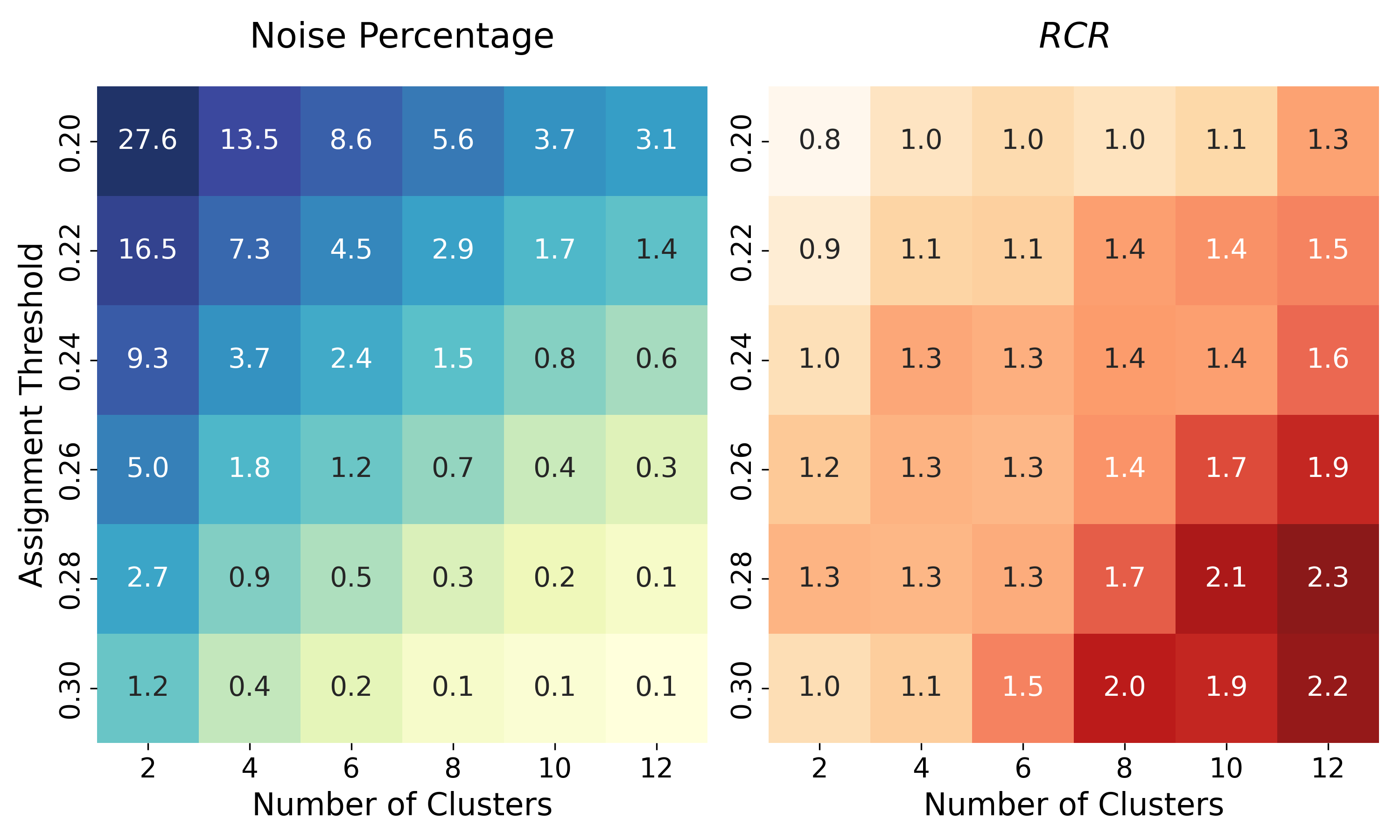}
    \caption{Noise percentages and $RCR$ at various cluster sizes and assignment thresholds. Colour scales are $\log_{10}$-normalised.}
    \label{fig:mse_noise}
\end{figure}

\begin{figure}
    \centering
    \includegraphics[width=0.95\linewidth]{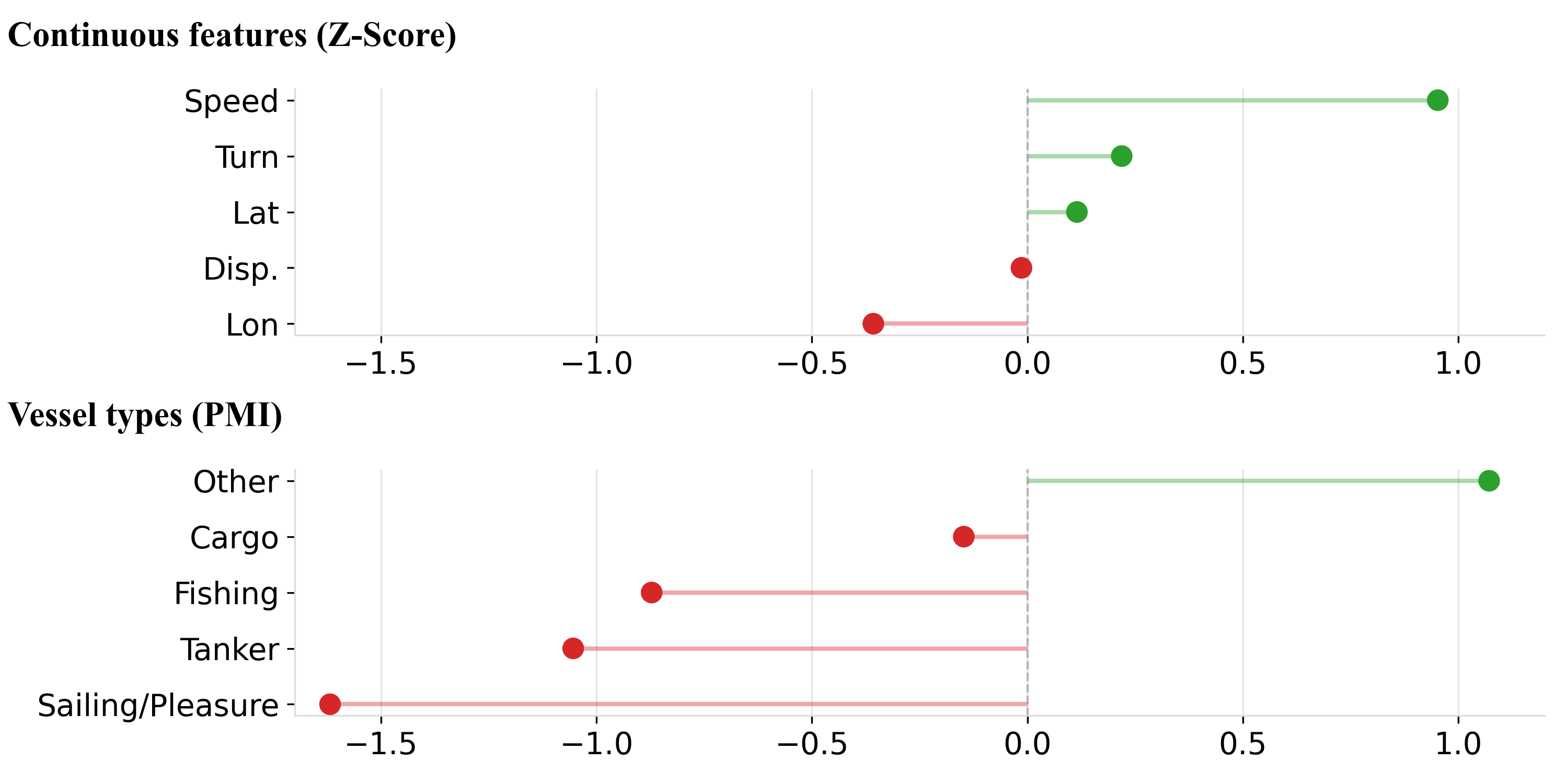}
    \caption{Z-score and $PMI$ for physical features for noise labelled points at 12 clusters with assignment threshold of 0.22. The two scores can be regarded as of similar scales and can be confidently compared.}
    \label{fig:noise_features}
\end{figure}

\subsection{Anomaly detection}
AC and CURE are applied with intrinsic noise detection to identify anomalous trajectories in the embedding space. The procedure is controlled by two task-specific hyperparameters: the number of clusters and the assignment threshold. Trajectories classified as noise exhibit higher reconstruction error, as the encoder is optimized on frequently occurring trajectory patterns, resulting in larger mean squared error (MSE) values for noise-labelled trajectories than for trajectories assigned to clusters.

To quantify this separation, a Reconstruction Contrast Ratio (RCR) is defined as
\[
\mathrm{RCR} = \frac{\mathbb{E}[\mathrm{MSE}_{\mathrm{noise}}]}{\mathbb{E}[\mathrm{MSE}_{\mathrm{cluster}}]},
\]
where $\mathbb{E}[\mathrm{MSE}_{\mathrm{noise}}]$ and $\mathbb{E}[\mathrm{MSE}_{\mathrm{cluster}}]$ denote the average reconstruction error of trajectories classified as noise and as cluster members, respectively. Figure~\ref{fig:mse_noise} illustrates the trade-off between noise proportion and RCR as a function of cluster count and assignment threshold. An assignment threshold of 0.22 with 12 clusters is used in subsequent analyses, corresponding to 1.4\% noise and an RCR of 1.5.

Assessing the demographic features characterising the noise-labelled trajectories on Figure \ref{fig:noise_features} we see that noise points at this setting tend to exhibit high speed and are situated at lower longitudes than cluster points. Additionally, noise trajectory are twice as likely than cluster points to be marked with label types other than Cargo, Fishing, Tanker, Sailing or Pleasure, with an especially low representation of Sailing/Pleasure vessels in noise trajectories, suggesting a high degree of cluster purity for recreational vessels.

\section{Conclusion}
This study presents a framework that combines BERT encoders with hierarchical clustering to effectively model and monitor large-scale maritime traffic. Processing 1.2 terabytes of AIS data produced physically interpretable clusters capturing distinct behaviors and fishing activities. Anomaly detection identified irregular trajectories in noisy, real-world maritime data.

All code used in this study is available \href{https://github.com/bertramhage/stcad}{here}.

\small
\bibliographystyle{IEEEtranN}
\bibliography{references}

\end{document}